%% file: iclr2027_conference.tex
\documentclass{article} 
\usepackage{iclr2027_conference,times}

\input{math_commands.tex}

\usepackage{hyperref}
\usepackage{url}
\usepackage{graphicx}
\usepackage{multirow}
\usepackage{booktabs}
\usepackage{makecell}
\usepackage{natbib}
\usepackage{xspace}
\usepackage[table]{xcolor}
\usepackage{enumitem}
\usepackage{amsmath}
\usepackage{mathtools}
\usepackage{amssymb}
\usepackage[most]{tcolorbox}
\usepackage[dvipsnames]{xcolor}
\usepackage{pifont}
\usepackage{url}
\usepackage{wrapfig}
\usepackage{caption}

\title{\raisebox{-0.2\height}{\includegraphics[height=1.15em]{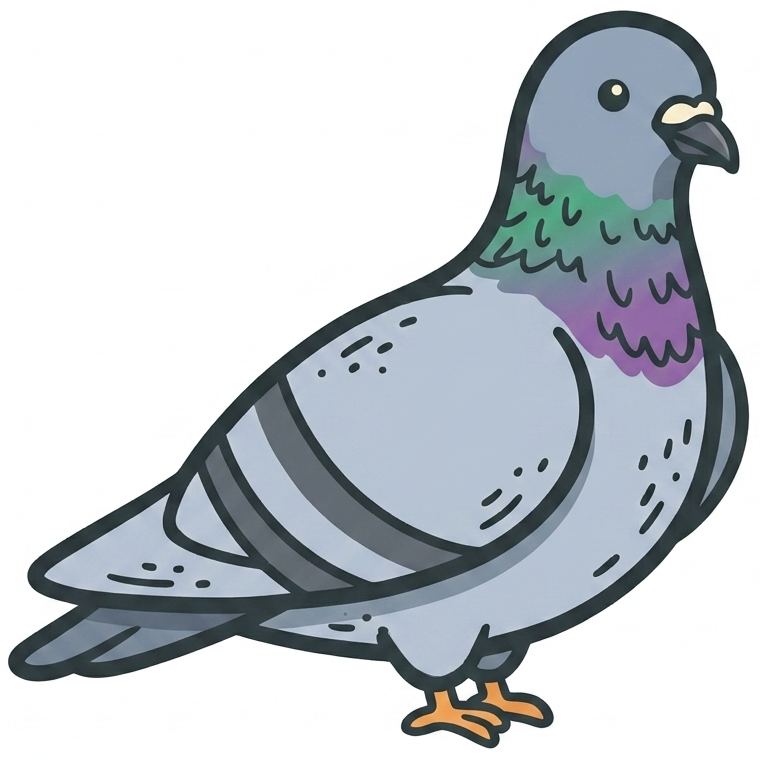}}\hspace{0.25em}PIGEONS: Incremental Long-Video Captioning with \underline{P}ersistent \underline{I}dentity \underline{G}allery and \underline{E}volving \underline{O}verarching \underline{N}arrative \underline{S}ummary}

\author{Xinlong Chen\textsuperscript{\textmd{1,2,3}}\thanks{This work was conducted during the author's internship at Kling Team.},
Weihong Lin\textsuperscript{\textmd{3}},
Yuanxing Zhang\textsuperscript{\textmd{3}},
Xinyu Liu\textsuperscript{\textmd{4}},
Yue Ding\textsuperscript{\textmd{1,2}},
Bozhou Li\textsuperscript{\textmd{3}}, \\
\textbf{Bohan Zeng}\textsuperscript{\textmd{5}}, 
\textbf{Yang Shi}\textsuperscript{\textmd{5}}, 
\textbf{Qiang Liu}\textsuperscript{\textmd{1,2}\thanks{Corresponding author: qiang.liu@nlpr.ia.ac.cn}},
\textbf{Pengfei Wan}\textsuperscript{\textmd{3}},
\textbf{Liang Wang}\textsuperscript{\textmd{1,2}},
\textbf{Tieniu Tan}\textsuperscript{\textmd{1,2,6}} \\
\textsuperscript{\textmd{1}}New Laboratory of Pattern Recognition (NLPR), \\ Institute of Automation, Chinese Academy of Sciences (CASIA) \\
\textsuperscript{\textmd{2}}School of Artificial Intelligence, University of Chinese Academy of Sciences \\
\textsuperscript{\textmd{3}}Kling Team \quad
\textsuperscript{\textmd{4}}The Hong Kong University of Science and Technology \\
\textsuperscript{\textmd{5}}Peking University \quad
\textsuperscript{\textmd{6}}Nanjing University \\ \\
\textbf{\texttt{Project webpage:~\url{https://pigeons-captioner.github.io/}}}
}

\newcommand{\circnum}[1]{\raisebox{-0.15ex}{\large\ding{#1}}}

\iclrfinalcopy 
\begin{document}

\maketitle
\begin{abstract}
Detailed video captioning has become an essential semantic interface for multimodal understanding and generation. While existing methods mainly target short videos, extending them to long videos remains challenging, primarily due to limited context windows of current video understanding models. A naive solution is to adopt segment-wise captioning followed by concatenation, yet this paradigm often suffers from inconsistent subject references and fragmented long-range narratives. To mitigate these issues, we introduce \textbf{PIGEONS}, an incremental long-video captioning model built upon a \underline{P}ersistent \underline{I}dentity \underline{G}allery and an \underline{E}volving \underline{O}verarching \underline{N}arrative \underline{S}ummary. Before caption generation, PIGEONS scans the entire video to construct an identity gallery that captures key subjects along with their visual references, thereby providing a stable referential anchor. The model then performs clip-by-clip captioning conditioned on the persistent gallery and an evolving overarching narrative summary, which compactly maintains and updates essential long-range contextual information after each captioning step. To enable these capabilities, we develop a three-stage training strategy that integrates supervised fine-tuning and reinforcement learning. Extensive experiments demonstrate that PIGEONS markedly improves long-video captioning performance, achieving superior accuracy, completeness, and subject referential consistency over existing methods, while maintaining competitive performance on short-video captioning.
\end{abstract}

\section{Introduction}
In the era of multimodal large language models (MLLMs), detailed video captioning has emerged as a crucial semantic interface that bridges complex perceptual signals with fine-grained linguistic representations~\citep{chai2025auroracap, chen2024sharegpt4video, chen2026avocado, yao2026timechat}. High-quality detailed video captions not only facilitate effective alignment among audio, visual, and textual modalities during pre-training~\citep{xu2025qwen3, team2025longcat}, but also provide rich semantic knowledge for downstream tasks~\citep{long2026seeing, du2025vc4vg}. Recent studies have consistently demonstrated that enhancing the quality of detailed video captions leads to notable performance gains across diverse multimodal applications~\citep{an2026onestory, team2026script, zhou2026aura, liu2026capability, shi2025mavors, hua2026vabench, liu2026rebind}.

Despite these advancements, existing research primarily focuses on relatively short videos~\citep{wang2024tarsier, yuan2025tarsier2, xu2026carebench, zhang2026vcapsbench, chen2025vidcapbench}, leaving long-form videos, which more closely resemble real-world scenarios, largely underexplored. A key challenge lies in the limited context windows of current video understanding models, which make it intractable to process an entire long video in a single pass while preserving sufficient temporal and spatial resolution for fine-grained understanding. To address this, a straightforward approach is segment-wise captioning, where a long video is first divided into shorter clips, each clip is then captioned independently, and the resulting clip-level captions are concatenated to form a complete video-level description.

\begin{figure}[t]
    \centering
    \includegraphics[width=\linewidth]{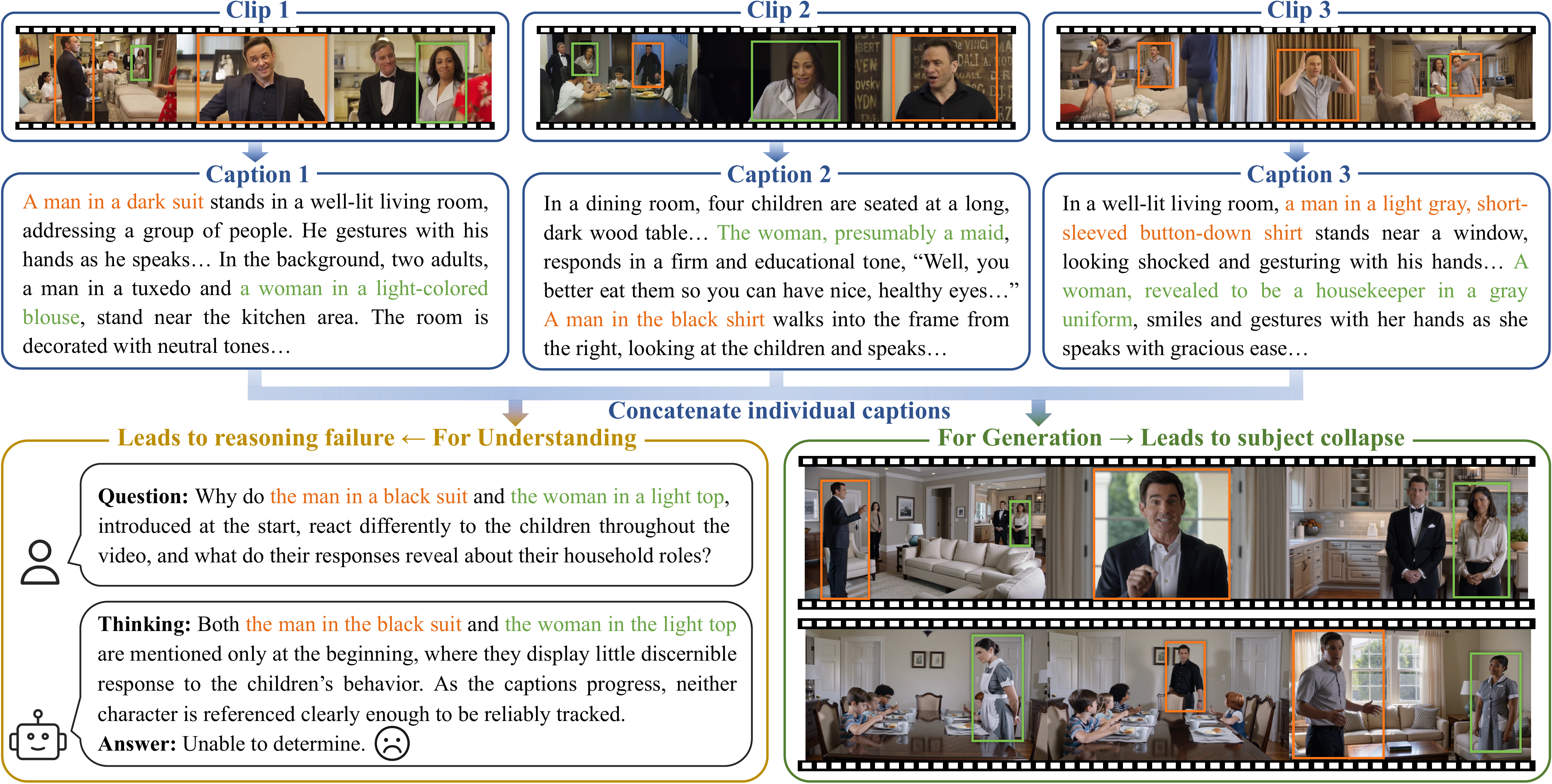}
    \caption{Downstream impacts of inconsistent subject references in naive segment-wise captioning. In understanding tasks, using such captions as the memory of an LLM agent may induce reasoning failures due to identity ambiguity. In generation tasks, leveraging these captions for video reconstruction could lead to subject collapse. Identical colors represent the same ground-truth subject.}
    \label{fig:teaser}
\end{figure}

However, this naive segment-wise approach introduces two critical limitations: (1) \textit{loss of subject referential consistency}, and (2) \textit{fragmented long-range narratives}. First, the same subject may be referred to using varying descriptions across clips. While locally accurate, the lack of stable identity associations makes it difficult to resolve coreferences in the final concatenated caption, thus confusing character relationships, disrupting narrative flow, and ultimately degrading the performance of downstream caption-based understanding and generation tasks, as shown in Figure~\ref{fig:teaser}. Second, isolated clip captioning ignores historical context. Models may redundantly reintroduce previously observed subjects or scenes, overlook established facts, or treat causally connected events as isolated local actions. While a straightforward remedy is to append all preceding captions as context, this approach scales poorly. As the video progresses, the accumulated caption history grows rapidly, which not only incurs substantial computational overhead but also introduces irrelevant details that distract the model from accurately describing the current event.

To mitigate these limitations, we introduce \textbf{PIGEONS}, an incremental long-video captioning model designed around a \underline{P}ersistent \underline{I}dentity \underline{G}allery and an \underline{E}volving \underline{O}verarching \underline{N}arrative \underline{S}ummary. Specifically, prior to segment-wise captioning, PIGEONS scans the entire video to construct a global identity gallery, assigning unique identifiers and reference images to key subjects. This gallery remains fixed throughout the subsequent clip-by-clip captioning process, serving as a stable anchor for global subject referential consistency. After captioning each clip, PIGEONS simultaneously extracts essential elements for subsequent narrative understanding (e.g., existing scenes, character dynamics, subject roles, and established facts) and integrates them into an evolving overarching narrative summary. While this summary captures long-range narrative context compactly, the caption of the immediately preceding clip is retained to ensure local continuity. Consequently, when generating the caption for the $n$-th clip ($n>1$), PIGEONS leverages three complementary sources of context: the persistent identity gallery for subject referential consistency, the evolving overarching summary for long-range narrative coherence, and the caption of the $(n-1) $-th clip for local continuity.

Equipping the model with these capabilities poses a dual challenge: identity-gallery construction requires precise spatiotemporal localization to identify representative subject images, while long-video captioning demands a comprehensive understanding of both video content and multimodal context. To tackle this, we design a tailored three-stage training strategy. Stage~1 applies supervised fine-tuning (SFT) on 16K identity-gallery construction samples, enabling preliminary capabilities in subject discovery and reference-image extraction. Stage~2 employs reinforcement learning (RL) with specialized reward functions to boost both gallery accuracy and construction efficiency. Stage~3 conducts SFT on 16K long-video captioning samples, training the model to generate multi-context-conditioned captions and dynamically update the evolving overarching summary. To mitigate catastrophic forgetting, identity-gallery construction data is mixed into this final stage. By jointly optimizing identity-gallery construction and context-aware captioning, PIGEONS is able to produce long-range cohesive captions with robust subject referential consistency.

Experimental results reveal that PIGEONS significantly outperforms existing models in long-video captioning, achieving superior accuracy, completeness, and subject referential consistency. Compared with naive segment-wise captioning, PIGEONS produces more coherent captions with stronger referential consistency, higher information density, and smoother cross-clip transitions. In addition, PIGEONS achieves highly competitive performance in short-video captioning scenarios that do not require segment-wise processing. Our contributions are summarized as follows:

\begin{itemize}[leftmargin=*]
    \item We introduce PIGEONS, an incremental long-video captioning model that leverages a persistent identity gallery and an evolving overarching narrative summary to ensure robust subject referential consistency and long-range narrative coherence.
    \item We develop a three-stage training strategy combining SFT and RL to jointly optimize precise identity-gallery construction and context-aware long-video captioning capabilities.
    \item Extensive experiments demonstrate that PIGEONS outperforms all existing models in terms of accuracy, completeness, and subject referential consistency for long-video captioning, while also delivering competitive performance in short-video captioning tasks.
\end{itemize}

\section{Related Work}
\subsection{Detailed Video Captioning with MLLMs}
Recent advances in MLLMs have propelled detailed video captioning. Early efforts focus on short, visual-only clips spanning tens of seconds, emphasizing fine-grained coverage and accuracy~\citep{wang2024tarsier, yuan2025tarsier2, chai2025auroracap, zhong2026owlcap}. Subsequent works incorporate auditory signals and extend target durations to the minute scale, enabling richer multimodal descriptions~\citep{tang2025video, wu2025ugc, geng2025longvale, chen2026avocado, ma2026omni, li2026towards}. Nevertheless, these time horizons still fall short of real-world long-form videos.

To handle long videos, existing research typically adopts a hierarchical paradigm~\citep{islam2024video, wei2025longcaptioning, chu2025fine}, decomposing videos into short clips for independent captioning, followed by global summarization. However, such isolated processing ignores cross-clip context, inevitably causing inconsistent subject references and fragmented narratives. Several works have attempted to alleviate these issues. StoryTeller~\citep{he2024storyteller} uses external movie cast information to assist subject identification across clips, but such auxiliary annotations are rarely available for general videos, and the simply concatenated captions still lack strong narrative coherence. RICE~\citep{yang2025addressing} explores subject association by prompting models to describe subject features, enabling language-based subject matching across captions. However, such textual association lacks stable visual anchors to reliably maintain referential consistency throughout long videos.

Distinct from these methods, PIGEONS explicitly maintains both subject identity and narrative context during incremental long-video captioning. By establishing a persistent visual identity gallery as a stable reference anchor, and continuously updating an overarching summary to preserve historical context, PIGEONS ensures consistent subject references and coherent long-range storytelling.

\subsection{Reinforcement Learning for Video Captioning}
Reinforcement learning~\citep{schulman2017proximal, shao2024deepseekmath, Zheng2025GroupSP, yu2026dapo} has been widely applied to video captioning with diverse reward designs. For instance, VideoChat-R1~\citep{li2025videochat} leverages event-recall rewards to improve caption quality, while VidBridge-R1~\citep{chen2026vidbridge} balances accuracy and completeness through a tailored reward mechanism. VideoCap-R1~\citep{meng2025videocap} decomposes captioning into structured thinking and generation stages, integrating dual scorers to elevate output quality. CapRL++~\citep{yang2026caprl++} uses the performance of an LLM on video-related question answering based on generated captions as an indirect reward signal for caption optimization. OmniScript~\citep{pu2026omniscript} optimizes long-form cinematic script generation using a temporally segmented reward that evaluates event-level accuracy across multiple dimensions.

Different from these approaches that directly optimize specific aspects of textual quality, PIGEONS targets the challenge of subject referential consistency in segmented long-video captioning. Instead of refining caption generation itself, PIGEONS leverages RL to optimize the upstream construction of the identity gallery, producing more accurate identity representations that provide explicit visual anchors to maintain subject referential consistency during subsequent captioning process.

\section{PIGEONS}\label{sec:pigeons}
\subsection{Overview}
To mitigate inconsistent subject referencing and fragmented long-range narratives caused by naive segment-wise captioning, we introduce \textbf{PIGEONS}. As illustrated in Figure~\ref{fig:procedure}, PIGEONS enhances the standard segment-wise captioning paradigm by integrating two key components: a \textit{persistent identity gallery} that provides visual subject reference anchors, and an \textit{evolving overarching narrative summary} that offers compact global narrative context. To enable these capabilities, we further develop a tailored three-stage training strategy combining SFT and RL. Details are provided below.

\subsection{Persistent Identity Gallery}
To maintain subject referential consistency across video clips, PIGEONS constructs a global \textit{identity gallery} prior to caption generation. This gallery stores unique identifiers and reference images for key subjects, and remains fixed throughout the subsequent captioning process to serve as a stable visual anchor for consistent subject referencing.

\begin{figure}[t]
    \centering
    \includegraphics[width=\linewidth]{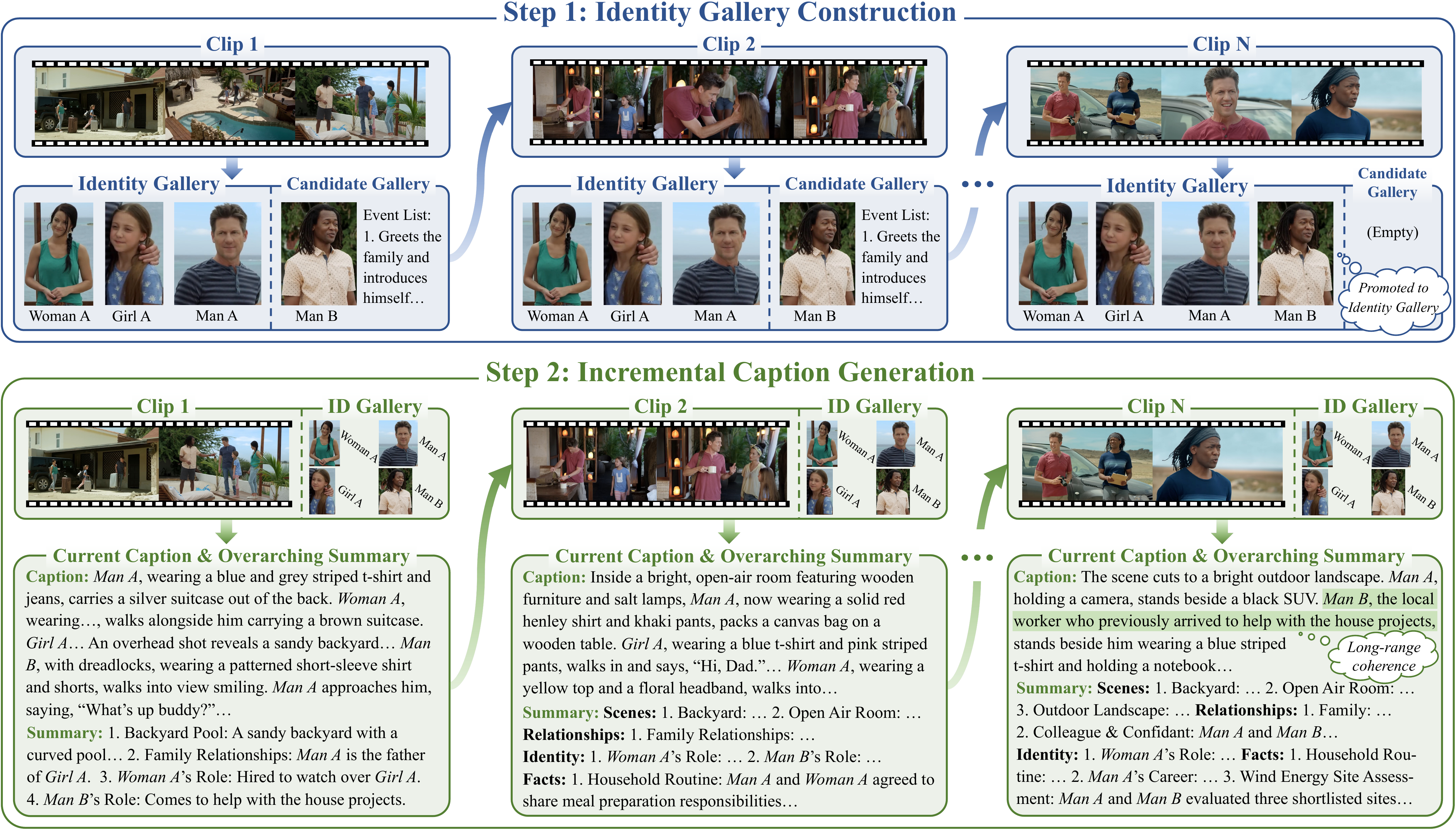}
    \caption{Inference pipeline of PIGEONS. The model first scans all video clips to construct a global identity gallery. Guided by this gallery, it sequentially generates captions for each clip, with consistent subject references in \textit{italic}. Concurrently, an overarching narrative summary is dynamically updated to help enhance long-range coherence and information density of the generated captions.}
    \label{fig:procedure}
\end{figure}

We define a \textit{subject} as an entity (not limited to human characters) that drives the plot and plays a significant role in the video's narrative. Recognizing that an entity may appear sparingly in early clips but become crucial later, we maintain an auxiliary \textit{candidate gallery} alongside the identity gallery. The candidate gallery tracks potential subjects in early stages along with their associated key events. Once an entity's key-event count reaches a predefined threshold (empirically set to 4), it is promoted to the identity gallery, as illustrated by \textit{Man~B} in the upper portion of Figure~\ref{fig:procedure}.

Formally, for a video $V$ partitioned into $N$ sequential clips $\{v_1, v_2, \dots, v_N\}$, let $\mathcal{G}_I^k$ and $\mathcal{G}_C^k$ represent the identity gallery and candidate gallery obtained after processing up to the $k$-th clip $v_k$, respectively. Initially, both galleries are empty, i.e., $\mathcal{G}_I^0 = \emptyset$ and $\mathcal{G}_C^0 = \emptyset$. At the $k$-th step, the model $\mathcal{M}$ takes the current clip and existing galleries as inputs to yield the updated galleries:
\begin{equation}
(\mathcal{G}_I^k, \mathcal{G}_C^k) = \mathcal{M}(v_k, \mathcal{G}_I^{k-1}, \mathcal{G}_C^{k-1}).
\end{equation}
After processing all $N$ clips, the final identity gallery $\mathcal{G}_I^N$ is fixed and serves as a persistent global visual anchor to ensure consistent subject referencing throughout the subsequent captioning process.

\subsection{Evolving Overarching Narrative Summary}
To leverage historical context without incurring context window explosion, PIGEONS maintains an \textit{evolving overarching narrative summary} that is updated after generating each clip's caption. Rather than naively accumulating verbose past captions, the model compresses the global narrative across four critical dimensions: \textit{existing scenes}, \textit{character relationships}, \textit{subject roles}, and \textit{established facts}. This multidimensional summarization preserves essential high-level semantic cues for understanding subsequent clips, ultimately yielding captions with enhanced long-range narrative coherence and higher information density, as depicted in the lower half of Figure~\ref{fig:procedure}.

While the evolving summary provides global context, the caption of the immediately preceding clip is additionally retained to maintain local narrative fluency. Formally, let $c_k$ and $s_k$ denote the generated caption and the updated narrative summary for clip $v_k$, with $c_0$ and $s_0$ initialized as empty strings. For the $k$-th clip, the model $\mathcal{M}$ generates its caption and updates the summary conditioned on the global identity gallery $\mathcal{G}_I^N$, the current video clip $v_k$, the preceding caption $c_{k-1}$, and the previous summary $s_{k-1}$:
\begin{equation}
(c_k, s_k) = \mathcal{M}(\mathcal{G}_I^N, v_k, c_{k-1}, s_{k-1}).
\end{equation}
The final long-video caption $c_{\text{video}}$ is then obtained by concatenating all clip-level captions:
\begin{equation}
c_{\text{video}} = \text{Concat}(c_1, c_2, \dots, c_N).
\end{equation}

\subsection{Three-Stage Training}
Building a reliable identity gallery requires precise spatiotemporal localization to identify representative subject images, while long-video captioning demands a comprehensive understanding of both video content and multimodal context. To equip our base model, Qwen3-Omni-Instruct~\citep{xu2025qwen3}, with these dual capabilities, we design a tailored three-stage training strategy.

\subsubsection{Stage~1: Identity Gallery SFT (IG-SFT)}
\begin{figure}[t]
    \centering
    \includegraphics[width=\linewidth]{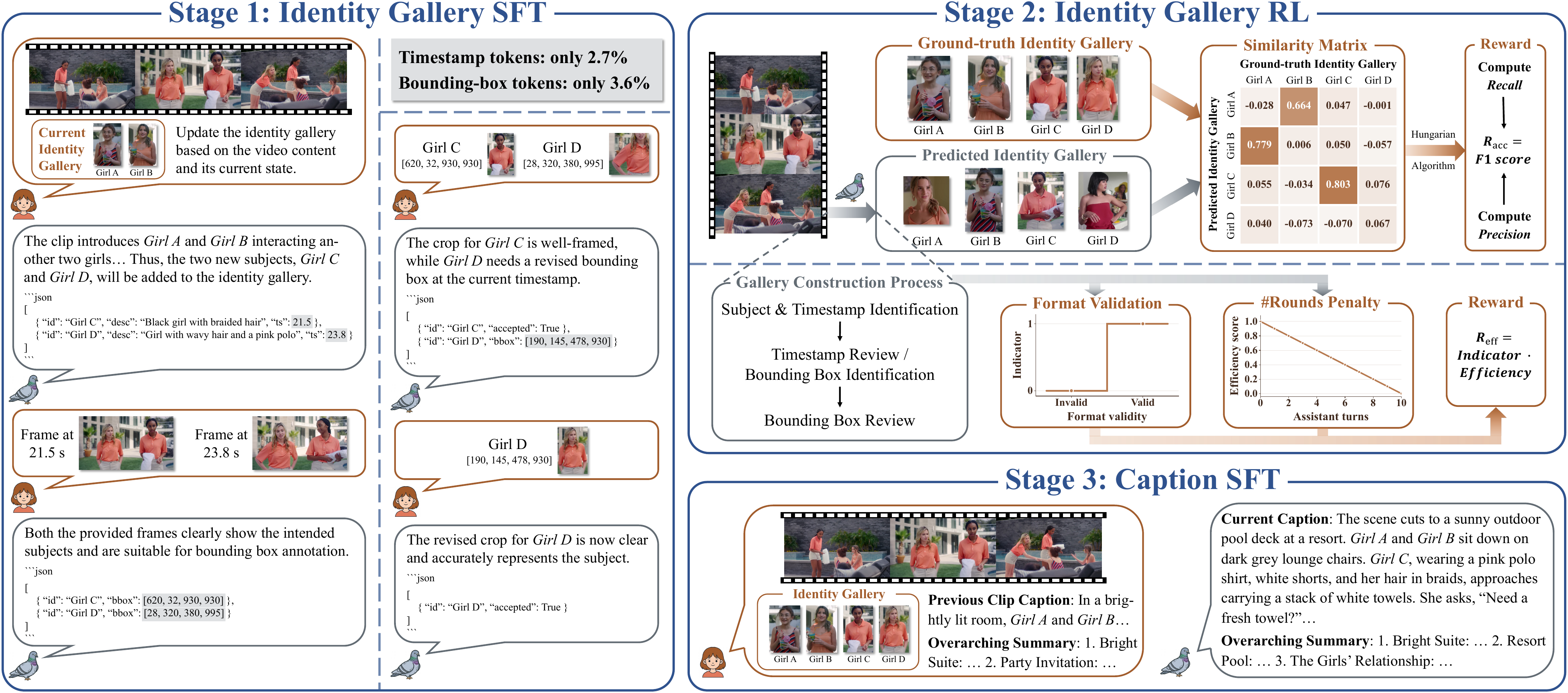}
    \caption{Overview of the three-stage training strategy. Stage~1 conducts SFT for identity gallery construction with self-review and refinement. Stage~2 further optimizes gallery construction via RL, improving both accuracy ($R_{\text{acc}}$) and efficiency ($R_{\text{eff}}$). Stage~3 performs caption SFT, enabling detailed caption generation conditioned on multiple contexts while evolving the overarching narrative summary. For clarity, only one clip from a full video is shown and the candidate gallery is omitted.}
    \label{fig:train}
\end{figure}

In the first stage, we curate a dataset of 16K identity gallery construction samples and perform SFT to teach the model to identify subjects in videos and determine the appropriate timestamps and spatial bounding boxes that capture discriminative and representative views of each subject. Considering that even minor temporal or spatial deviations can substantially degrade the quality of reference images, we explicitly incorporate a verification and refinement procedure, encouraging the model to iteratively self-evaluate and correct low-quality crops, thereby producing higher-quality identity galleries, as illustrated in the left part of Figure~\ref{fig:train}.

\subsubsection{Stage~2: Identity Gallery RL (IG-RL)}
During identity gallery construction, the critical timestamp and bounding box information occupies only 2.7\% and 3.6\% of the total output sequence, making token-level SFT insufficient for providing dense supervision over these key decisions. Therefore, in the second stage, we introduce RL, specifically Group Sequence Policy Optimization (GSPO)~\citep{Zheng2025GroupSP}, to further optimize gallery construction. As illustrated in the upper right part of Figure~\ref{fig:train}, we design two outcome-oriented reward functions targeting gallery quality and construction efficiency, respectively, and apply RL on a randomly sampled subset of 2K instances from the preceding SFT stage.

\textbf{Identity Gallery Accuracy Reward $R_{\text{acc}}$.} Since the construction process of an identity gallery is not uniquely determined, we define $R_{\text{acc}}$ based on the similarity between the predicted and the ground-truth (GT) identity galleries. Formally, for a video segmented into $N$ clips, the model ultimately constructs an identity gallery $\mathcal{G}_I^N = \{(s_p, r_p)\}_{p=1}^{P}$ consisting of $P$ subject-reference pairs, and the corresponding GT identity gallery is denoted as $\tilde{\mathcal{G}}_I = \{(\tilde{s}_q, \tilde{r}_q)\}_{q=1}^{Q}$, containing $Q$ pairs, where $s$ denotes the subject identifier and $r$ represents the corresponding reference image.

To measure the similarity between two galleries, we first define the pairwise similarity between their entries $(s_p, r_p)$ and $(\tilde{s}_q, \tilde{r}_q)$. Since subject identifier $s$ is merely a specific instantiation without intrinsic semantic correspondence, while the reference image $r$ directly determines the quality of identity representation, we measure entry-level similarity based solely on the reference images. Furthermore, as an identity gallery is inherently an unordered set, we formulate the gallery-level similarity as a maximum-weight bipartite matching problem with a one-to-one assignment constraint.

Specifically, we first employ the face detection confidence score $\text{Conf}(\cdot)$ from ArcFace~\citep{deng2019arcface} to determine whether a reference image contains a recognizable face. For two face-containing images, we compute their similarity using ArcFace. For two non-face images, we use DINOv3~\citep{simeoni2025dinov3} to measure their visual similarity. The similarity between a face-containing image and a non-face image is set to zero. Formally:
\begin{equation}
\text{Sim}(r_p, \tilde{r}_q) =  \begin{cases}  \text{ArcFace}(r_p, \tilde{r}_q), & \text{if } \text{Conf}(r_p) > \gamma \text{ and } \text{Conf}(\tilde{r}_q) > \gamma, \\ \text{DINOv3}(r_p, \tilde{r}_q), & \text{if } \text{Conf}(r_p) \leq \gamma \text{ and } \text{Conf}(\tilde{r}_q) \leq \gamma, \\ 0, & \text{otherwise}, \end{cases}
\end{equation}
where the face confidence threshold $\gamma$ is empirically set to $0.35$.

Based on the pairwise similarities, we build a similarity matrix $S \in \mathbb{R}^{P \times Q}$ between the predicted and GT galleries. To avoid unreliable assignments caused by weakly correlated pairs, we apply metric-specific thresholds: $S_{pq}$ is set to $\text{Sim}(r_p, \tilde{r}_q)$ only if the similarity exceeds $\tau_{\text{arc}} = 0.35$ for face pairs or $\tau_{\text{dino}} = 0.55$ for non-face pairs; otherwise, $S_{pq}$ is set to zero. The Hungarian algorithm~\citep{Hungarian} is then applied to find the globally optimal bipartite matching, and the total similarity score $S_{\text{match}}$ is given by:
\begin{equation}
S_{\text{match}} = \max_{\mathcal{H}} \sum_{(p,q) \in \mathcal{H}} S_{pq},
\end{equation}
where $\mathcal{H}$ denotes a valid bipartite matching under the one-to-one assignment constraint.

From a physical interpretation, $S_{\text{match}}$ represents the quality of the reference images in $\mathcal{G}_I^N$, which takes values in the range $[0, \text{min}(P, Q)]$. The precision and recall are then computed as:
\begin{equation}
\text{Pre} = S_{\text{match}} \big{/} P, \qquad \text{Rec} = S_{\text{match}} \big{/} Q.
\end{equation}
Ultimately, the overall matching quality between the two identity galleries is quantified by the $F_1$ score, which serves as the identity gallery accuracy reward:
\begin{equation}
R_{\text{acc}} = 2 \cdot \text{Pre} \cdot \text{Rec} \big{/} {(\text{Pre} + \text{Rec})}.
\end{equation}
\textbf{Efficiency-Aware Format Reward $R_{\text{eff}}$.} To enforce structured outputs and improve construction efficiency by avoiding excessive self-reviews, we design an efficiency-aware format reward, $R_{\text{eff}}$.

Specifically, we first verify whether the output satisfies the predefined format constraints, such as valid JSON syntax and required key fields. Invalid outputs receive zero reward, while valid ones are penalized linearly by the number of interaction rounds $k$ to discourage unnecessary iterations:
\begin{equation}
R_{\text{eff}}=\mathbb{I}_{\text{valid}} \cdot (1-\alpha k),
\end{equation}
where the indicator $\mathbb{I}_{\text{valid}} \in \{0, 1\}$ denotes format validity, and the penalty coefficient $\alpha$ is set to $0.1$.

Given the primary importance of gallery quality, the final reward $R$ is defined as a weighted combination of the accuracy reward and the efficiency-aware format reward:
\begin{equation}
R = R_{\text{acc}} + \lambda R_{\text{eff}},
\end{equation}
where the weight coefficient $\lambda$ is set to $0.2$.

\subsubsection{Stage~3: Caption SFT (Cap-SFT)} 
After learning identity gallery construction, we further perform SFT on 16K long-video captioning samples in the third stage, to enhance the model's ability to generate detailed descriptions conditioned on the identity gallery, the evolving summary, and the preceding caption, while simultaneously updating the overarching narrative summary to facilitate the understanding of subsequent clips, as illustrated in the lower right part of Figure~\ref{fig:train}. To mitigate catastrophic forgetting, identity gallery construction samples are mixed into this final training stage.

By jointly optimizing identity gallery construction and multi-context-conditioned long-video captioning, PIGEONS effectively improves its ability to generate long-range cohesive captions with robust subject referential consistency. Detailed data construction procedures and additional training details are provided in Appendix~\ref{app:training_details}.

\begin{table}[t]
    \centering
    \caption{Model performance on CapRiCorn-1K, which evaluates overall caption quality (Acc \& Cov) and subject referential consistency (Ref) across temporal scales.}
    \resizebox{\linewidth}{!}{
    \begin{tabular}{l|>{\columncolor{gray!10}}c>{\columncolor{gray!10}}c>{\columncolor{gray!10}}c|ccc|ccc|ccc|ccc}
    \toprule
    \multirow{2}{*}{\textbf{Model}} & \multicolumn{3}{>{\columncolor{gray!10}}c|}{\textbf{Overall}} & \multicolumn{3}{c|}{\textbf{(0, 2]~min}} & \multicolumn{3}{c|}{\textbf{(2, 5]~min}} & \multicolumn{3}{c|}{\textbf{(5, 8]~min}} & \multicolumn{3}{c}{\textbf{(8, 10]~min}} \\
    ~ & \textbf{Acc} & \textbf{Cov} & \textbf{Ref} & \textbf{Acc} & \textbf{Cov} & \textbf{Ref} & \textbf{Acc} & \textbf{Cov} & \textbf{Ref} & \textbf{Acc} & \textbf{Cov} & \textbf{Ref} & \textbf{Acc} & \textbf{Cov} & \textbf{Ref} \\
    \midrule
    Gemini-3.1-Pro & \underline{42.5} & \underline{53.3} & 39.1 & 40.9 & 53.4 & 42.4 & \underline{44.7} & \underline{54.8} & \underline{40.4} & \underline{42.4} & \underline{52.7} & 35.3 & \underline{42.2} & \underline{51.6} & \underline{35.4} \\
    Gemini-3-Flash & 41.5 & 52.8 & \underline{39.6} & \underline{42.8} & 55.9 & \underline{46.3} & 42.1 & 52.9 & 38.1 & 41.1 & 51.3 & \underline{36.6} & 38.5 & 48.3 & 32.3 \\
    \midrule
    Qwen2.5-Omni & 5.1 & 13.2 & 0.6 & 6.7 & 17.4 & 1.2 & 5.9 & 13.7 & 0.5 & 3.4 & 10.0 & 0.3 & 2.6 & 8.5 & 0.1 \\
    OmniVinci & 5.9 & 13.3 & 1.2 & 9.9 & 21.2 & 2.5 & 5.3 & 12.0 & 0.6 & 3.2 & 8.3 & 0.4 & 2.5 & 6.2 & 0.5 \\
    ARC-Qwen-Video & 6.9 & 10.9 & 2.0 & 9.2 & 15.3 & 3.4 & 7.2 & 11.0 & 1.8 & 5.8 & 9.0 & 1.4 & 2.8 & 4.4 & 0.4 \\
    Qwen3-Omni-Instruct & 10.3 & 20.2 & 1.6 & 13.3 & 24.6 & 2.5 & 10.6 & 20.3 & 1.6 & 8.0 & 17.6 & 0.9 & 6.5 & 14.5 & 0.7 \\
    UGC-VideoCaptioner & 11.8 & 21.8 & 3.6 & 17.4 & 30.6 & 7.4 & 11.0 & 20.1 & 2.1 & 8.3 & 17.0 & 1.4 & 6.5 & 13.2 & 0.9 \\
    Qwen3-Omni-Captioner & 14.3 & 27.5 & 4.1 & 18.1 & 33.5 & 7.0 & 14.4 & 27.4 & 3.4 & 11.4 & 23.0 & 2.3 & 10.2 & 21.4 & 1.9 \\
    ASID-Captioner & 18.9 & 31.1 & 12.9 & 30.2 & 47.3 & 26.3 & 18.6 & 30.4 & 10.3 & 10.5 & 19.3 & 3.8 & 7.7 & 14.9 & 1.9 \\
    video-SALMONN-2 & 22.5 & 37.6 & 11.3 & 27.6 & 46.0 & 18.2 & 23.2 & 37.6 & 10.6 & 19.9 & 33.2 & 7.1 & 14.3 & 26.2 & 4.0 \\
    DiaDem & 24.6 & 35.8 & 14.5 & 40.0 & 54.7 & 31.3 & 23.1 & 33.2 & 10.2 & 13.6 & 23.0 & 3.3 & 10.3 & 18.3 & 2.0 \\
    AVoCaDO & 28.8 & 41.9 & 18.4 & \textbf{43.7} & \textbf{60.6} & 36.6 & 29.8 & 42.4 & 15.8 & 17.0 & 27.5 & 5.2 & 12.9 & 22.3 & 3.1 \\
    \midrule
    PIGEONS (Ours) & \textbf{50.7} & \textbf{65.4} & \textbf{51.8} & 42.5 & \underline{60.3} & \textbf{51.3} & \textbf{51.0} & \textbf{65.9} & \textbf{52.8} & \textbf{55.7} & \textbf{68.6} & \textbf{52.2} & \textbf{60.1} & \textbf{70.8} & \textbf{50.5} \\
    \bottomrule
    \end{tabular}
    }
    \label{tab:main1}
\end{table}

\section{Experiments}
\subsection{Experimental Settings}
We compare PIGEONS against a broad range of general-purpose video understanding models, including open-source models such as Qwen2.5-Omni~\citep{xu2025qwen2}, OmniVinci~\citep{ye2026omnivinci}, ARC-Qwen-Video~\citep{ge2025arc}, and the Qwen3-Omni series~\citep{xu2025qwen3, ma2026omni}, as well as commercial models including Gemini-3.1-Pro~\citep{gemini31pro} and Gemini-3-Flash~\citep{gemini3flash}. We further compare against specialized video captioning models, including video-SALMONN-2~\citep{tang2025video}, UGC-VideoCaptioner~\citep{wu2025ugc}, AVoCaDO~\citep{chen2026avocado}, DiaDem~\citep{chen2026diadem}, and ASID-Captioner~\citep{li2026towards}.

Evaluation is conducted on two video captioning benchmarks, CapRiCorn-1K~\citep{chen2026capricorn} and UGC-VideoCap~\citep{wu2025ugc}. Additionally, we indirectly assess caption quality on WorldSense~\citep{hong2026worldsense} by prompting a judge model to answer video-related questions based solely on the generated captions. Together, these benchmarks cover videos ranging from a few seconds to over 10 minutes, enabling evaluation across a wide range of video durations.

\begin{table}[t]
    \centering
    \caption{Model performance on UGC-VideoCap, which contains videos shorter than 60 seconds, and WorldSense, where caption quality is evaluated by prompting a judge model (Gemini-2.5-Pro~\citep{comanici2025gemini}, following AVoCaDO) to answer questions based on the generated captions.}
    \resizebox{\linewidth}{!}{
    \begin{tabular}{l|cccc|ccccc}
    \toprule
        \multirow{2}{*}{\textbf{Model}} & \multicolumn{4}{c|}{\textbf{UGC-VideoCap}} & \multicolumn{5}{c}{\textbf{WorldSense (Acc by Caption)}} \\
        ~ & \textbf{Audio} & \textbf{Visual} & \textbf{Detail} & \textbf{Avg} & \textbf{(0, 2] min} & \textbf{(2, 4] min} & \textbf{(4, 6] min} & \textbf{(6, 8] min} & \textbf{(8, 12] min} \\
        \midrule
        \textcolor{gray!100}{Gemini-3.1-Pro} & \textcolor{gray!100}{72.6} & \textcolor{gray!100}{77.8} & \textcolor{gray!100}{76.2} & \textcolor{gray!100}{75.6} & \textcolor{gray!100}{28.8} & \textcolor{gray!100}{27.9} & \textcolor{gray!100}{29.9} & \textcolor{gray!100}{26.5} & \textcolor{gray!100}{36.2} \\
        \textcolor{gray!100}{Gemini-3-Flash} & \textcolor{gray!100}{71.4} & \textcolor{gray!100}{76.7} & \textcolor{gray!100}{75.1} & \textcolor{gray!100}{74.4} & \textcolor{gray!100}{25.8} & \textcolor{gray!100}{24.7} & \textcolor{gray!100}{24.4} & \textcolor{gray!100}{23.5} & \textcolor{gray!100}{30.2} \\
        \midrule
        Qwen2.5-Omni & 46.9 & 66.1 & 60.0 & 57.7 & 10.5 & 6.6 & 4.5 & 3.7 & 4.0 \\
        UGC-VideoCaptioner & 61.4 & 58.4 & 57.5 & 59.1 & 12.9 & 8.9 & 6.3 & 8.8 & 7.4 \\
        video-SALMONN-2 & 61.8 & 71.4 & 68.5 & 67.2 & 18.4 & 17.3 & 17.2 & 17.7 & 22.2 \\
        Qwen3-Omni-Instruct & 67.5 & 74.8 & \underline{72.3} & 71.5 & 13.5 & 11.0 & 13.1 & 7.4 & 14.1 \\
        Qwen3-Omni-Captioner & \underline{69.0} & \underline{75.5} & \underline{72.3} & \underline{72.5} & 15.7 & 11.8 & 10.4 & 9.6 & 11.4 \\
        AVoCaDO & \textbf{73.0} & 74.6 & 71.8 & \textbf{73.2} & \textbf{27.4} & \underline{23.1} & \underline{20.4} & \underline{22.1} & \underline{25.5} \\
        \midrule
        PIGEONS (Ours) & 65.9 & \textbf{75.9} & \textbf{73.1} & 71.7 & \underline{25.1} & \textbf{25.8} & \textbf{24.9} & \textbf{30.2} & \textbf{35.6} \\
    \bottomrule
    \end{tabular}
    }
    \label{tab:main2}
\end{table}

\subsection{Experimental Results}
Table~\ref{tab:main1} presents the performance of various models on CapRiCorn-1K, which evaluates both overall caption quality and subject referential consistency across varying temporal scales. The results demonstrate that PIGEONS substantially outperforms existing models, including the closed-source Gemini series, achieving superior caption accuracy, coverage, and subject referential consistency averaged across all temporal scales. Moreover, powered by its persistent identity gallery and evolving overarching narrative summary, PIGEONS exhibits increasingly larger advantages as video duration grows. For videos shorter than two minutes that can be directly processed within the context window without segmentation (as detailed in Appendix~\ref{app:implement_details}), PIGEONS also achieves comparable overall caption quality to the previous state-of-the-art open-source model, AVoCaDO, while remarkably boosting subject referential consistency from 36.6 to 51.3.

Table~\ref{tab:main2} reports the evaluation results on UGC-VideoCap and WorldSense. UGC-VideoCap directly evaluates short-video (under 60 seconds) caption quality across audio, visual, and detail dimensions, while WorldSense provides an indirect evaluation by prompting a judge model to answer video-related questions based on generated captions. Consistent with the trends in Table~\ref{tab:main1}, PIGEONS demonstrates competitive performance with AVoCaDO on videos under two minutes, where segmentation is not required. For longer videos, PIGEONS substantially surpasses existing open-source counterparts, with the performance gap becoming increasingly significant as video duration extends.

\subsection{Ablation Studies of the Three-Stage Training Strategy}
\begin{wraptable}{r}{0.6\textwidth}
    \centering
    \vspace{-18pt}
    \caption{Ablation studies of the three-stage training strategy. ``Segment \& Concat'' refers to segment-wise captioning followed by concatenation. ``IG'' and ``Cap'' denote identity gallery and caption, respectively. Circled numbers indicate the actual training order of each stage.}
    \resizebox{\linewidth}{!}{
    \begin{tabular}{@{\hspace{2pt}}c@{\hspace{4pt}}|l@{\hspace{4pt}}c@{\hspace{6pt}}c@{\hspace{6pt}}c@{\hspace{6pt}}c|ccc}
    \toprule
        \multirow{2}{*}{\textbf{Row}} & \multirow{2}{*}{\textbf{Model}} & \multirow{2}{*}{\textbf{\makecell[c]{Segment\\\& Concat}}} & \multicolumn{3}{c|}{\textbf{Training Recipe}} & \multicolumn{3}{c}{\textbf{CapRiCorn-1K}} \\
        ~ & ~ & ~ & IG-SFT & IG-RL & Cap-SFT & Acc & Cov & Ref \\
        \midrule
        1 & \multirow{2}{*}{Qwen3-Omni-Ins} & \ding{55} & \textbf{--} & \textbf{--} & \textbf{--} & 10.3 & 20.2 & 1.6 \\
        2 & ~ & \ding{51} & \textbf{--} & \textbf{--} & \textbf{--} & 24.5 & 37.8 & 9.6 \\
        \midrule
        3 & \multirow{3}{*}{Qwen3-Omni-Ins} & \ding{51} & \multicolumn{3}{c|}{Cap-SFT w/o IG} & 49.1 & 64.4 & 28.1 \\
        4 & ~ & \ding{51} & \textbf{--} & \textbf{--} & \circnum{172} & 49.0 & 64.0 & 49.4 \\
        5 & ~ & \ding{51} & \textbf{--} & \circnum{173} & \circnum{172} & 47.4 & 62.6 & 48.4 \\
        \midrule
        6 & \multirow{2}{*}{Qwen3-Omni-Ins} & \ding{51} & \circnum{172} & \textbf{--} & \circnum{173} & 49.3 & 63.8 & 49.9 \\
        7 & ~ & \ding{51} & \circnum{172} & \circnum{174} & \circnum{173} & 46.9 & 62.1 & 48.7 \\
        \midrule
        8 & \makecell[l]{PIGEONS\\(IG-RL w/o $R_\text{eff}$)} & \ding{51} & \circnum{172} & \circnum{173} & \circnum{174} & 49.8 & 64.3 & 50.2 \\
        \midrule
        9 & PIGEONS (Ours) & \ding{51} & \circnum{172} & \circnum{173} & \circnum{174} & \textbf{50.7} & \textbf{65.4} & \textbf{51.8} \\
    \bottomrule
    \end{tabular}
    }
    \label{tab:abla_train}
\end{wraptable}
In Table~\ref{tab:abla_train}, we conduct an in-depth analysis of the individual components and specific training orders within our post-training pipeline.

Rows~1 and~2 demonstrate the efficacy of segment-wise captioning with concatenation compared to directly captioning the entire video using the base model. By providing richer audiovisual information, the segment-wise strategy substantially improves the overall caption quality.

Row~3 evaluates Cap-SFT without identity-gallery data and conditioning, where the model learns to generate captions based solely on the evolving overarching narrative summary and the preceding caption. Although this setting improves overall caption quality over naive segment-wise captioning (Row~2), it yields considerably lower subject referential consistency than subsequent ablations with identity-gallery conditioning, underscoring the necessity of explicit identity anchoring to maintain consistent subject references across video clips.

Rows~4 and~5 examine the effects of applying Cap-SFT alone, and appending IG-RL after Cap-SFT, respectively. By enabling the model to leverage identity-gallery information during caption generation, Cap-SFT alone achieves strong gains in subject referential consistency over Row~3. However, it still falls short of the fully trained PIGEONS due to the lack of a dedicated preliminary learning process for identity-gallery construction. Moreover, appending IG-RL after Cap-SFT leads to consistent degradation across all metrics, suggesting that optimizing identity-gallery construction after caption learning could disrupt the captioning capability acquired during Cap-SFT.

Rows~6 and~7 incorporate IG-SFT as the initial training stage. Compared to their counterparts without IG-SFT (Rows~4 and~5), the overall captioning metrics fluctuate slightly, while subject referential consistency shows modest gains. Nevertheless, they still underperform PIGEONS, indicating that identity-gallery learning is more effective when IG-SFT and IG-RL are executed consecutively prior to Cap-SFT, rather than skipping IG-RL or misplacing it after caption learning.

Taken together, these results validate both the necessity and the specific ordering of our three-stage training strategy: IG-SFT first establishes basic identity-gallery construction capability, IG-RL then refines this capability through targeted rewards, and Cap-SFT is placed last to directly optimize caption generation while leveraging the previously acquired identity-gallery construction skills.

Finally, Row~8 investigates the impact of removing the efficiency-aware format reward $R_\text{eff}$ from the IG-RL stage. Through qualitative analysis, we observe that without $R_\text{eff}$, the model exhibits weaker generalization on unseen samples and tends to engage in excessive verification-and-refinement loops. As a result, it frequently fails to complete identity-gallery construction within the limited interaction budget, which ultimately compromises the subsequent long-video captioning quality.

\subsection{Comparison with Segment-Wise Captioning and Concatenation}
\begin{wraptable}{r}{0.46\textwidth}
    \centering
    \vspace{-18pt}
    \caption{Comparison with AVoCaDO-based segment-wise captioning and concatenation.}
    \resizebox{\linewidth}{!}{
    \begin{tabular}{l|ccc|c}
    \toprule
        \multirow{2}{*}{\textbf{Model}} & \multicolumn{3}{c|}{\textbf{CapRiCorn-1K}} & \multirow{2}{*}{\textbf{\makecell[c]{Avg Caption\\Length}}} \\
        ~ & Acc & Cov & Ref \\
        \midrule
        AVoCaDO-Concat & 49.8 & 64.1 & 34.1 & 3025.7 \\
        PIGEONS (Ours) & \textbf{50.7} & \textbf{65.4} & \textbf{51.8} & \textbf{1271.5} \\
    \bottomrule
    \end{tabular}
    }
    \label{tab:abla_avocado_concat}
\end{wraptable}
To further demonstrate the advantages provided by the two key components of PIGEONS, namely the persistent identity gallery and the evolving overarching narrative summary, we compare PIGEONS with AVoCaDO, the leading baseline for videos within two minutes, under the similar segment-wise captioning and concatenation setting. As shown in Table~\ref{tab:abla_avocado_concat}, PIGEONS achieves higher captioning accuracy and coverage while reducing the average caption length by more than half, suggesting substantially higher information density. Moreover, it remarkably enhances subject referential consistency within the captions, further validating the effectiveness of our approach.

\section{Conclusion}
This work focuses on the important yet underexplored challenge of detailed long-video captioning. We first identify that the naive segment-wise captioning strategy suffers from two critical issues: inconsistent subject references across clips and degraded long-range narrative coherence. To mitigate these challenges, we introduce PIGEONS, an incremental long-video captioning model equipped with a persistent identity gallery and an evolving overarching narrative summary. The identity gallery provides stable visual anchors for consistent subject references, while the evolving summary efficiently preserves essential historical context for coherent long-range storytelling. 

To enable these capabilities, we develop a tailored three-stage training strategy that combines SFT and RL, progressively equipping the model with reliable identity-gallery construction and context-aware long-video captioning abilities. Extensive experiments demonstrate that PIGEONS substantially advances long-video captioning performance, achieving superior accuracy, completeness, and subject referential consistency compared to existing approaches. Meanwhile, PIGEONS maintains competitive performance in short-video scenarios where segmentation is unnecessary. Comprehensive ablation studies further verify the contribution of each component in our design and training pipeline, underscoring the effectiveness of our approach.

\clearpage

\bibliography{iclr2027_conference}
\bibliographystyle{iclr2027_conference}

\clearpage
\appendix
\section{Additional Training Details}\label{app:training_details}
\subsection{Training Data Construction}
To improve the model's generalization across diverse video domains and real-world application scenarios, we construct a training corpus by collecting and annotating videos from multiple datasets with complementary distributions. Specifically, we first select videos from movie-oriented datasets, including CinePile~\citep{rawal2024cinepile}, MovieStory101~\citep{he2024storyteller}, MovieChat-1K~\citep{song2024moviechat}, and LSMDC~\citep{rohrbach2017movie}, which typically exhibit coherent narrative structures and contain major characters or objects that recur consistently across shots. To cover a broader spectrum of open-domain, real-world audiovisual content, we further incorporate videos from TikTok-10M~\citep{tiktok_10m_2025}, encompassing vlogs, daily recordings, artistic performances, and short-form narratives. In addition, FineVideo~\citep{Farre2024FineVideo} is included to further enrich the diversity of thematic domains and stylistic distributions.

To accommodate the context-window limitation of the base model, we uniformly partition long videos into segments with a maximum duration of two minutes, while keeping shorter videos intact. Each resulting segment is subsequently annotated by Gemini-3.1-Pro for both identity gallery construction and caption generation.

The identity gallery construction consists of subject identification, iterative proposal and refinement of timestamps and bounding boxes for reference images, and the promotion of candidate entities into the final identity gallery. To ensure the quality and reliability of the constructed galleries, we implement a rigorous quality-control procedure powered by GPT-5.5. Specifically, this procedure first filters out reference images with low visual quality, including blurriness, misalignment, or indistinct visual features. Subsequently, it removes samples featuring false-positive subjects that contribute negligibly to the plot or overall video narrative.

For caption annotation, Gemini-3.1-Pro generates segment-level captions conditioned on the identity gallery, the evolving overarching narrative summary, and the caption of the preceding segment. Meanwhile, the overarching summary is updated dynamically with each newly generated caption to preserve long-range contextual coherence. We then apply a strict quality-control process to the captions, discarding samples with repetition collapses or instances where a gallery entity remains unmentioned across all segment captions within a video. Finally, GPT-5.5 is utilized to identify and exclude captions containing incorrect associations between subject identities and described events.

After the above annotation and quality-control procedures, our final corpus retains 9K videos from the movie-oriented datasets (i.e., CinePile, MovieStory101, MovieChat-1K, and LSMDC), 6K from TikTok-10M, and 1K from FineVideo. In aggregate, the resulting dataset contains 16K full-length videos with an average duration of 3.43 minutes, yielding a total of 37K video clips after segmentation for training. Detailed annotation prompts are provided in Appendix~\ref{app:prompts}.

\subsection{Group Sequence Policy Optimization}
While Group Relative Policy Optimization (GRPO)~\citep{shao2024deepseekmath} eliminates the need for a separate critic model by estimating advantages within sampled response groups, its token-level importance weighting often introduce high-variance optimization noise, particularly for long sequences. To better align the optimization objective with sequence-level rewards, we adopt Group Sequence Policy Optimization (GSPO)~\citep{Zheng2025GroupSP}. By replacing the token-level importance ratio in GRPO with a length-normalized sequence-level counterpart, and applying clipping over the entire response, GSPO facilitates more stable and efficient policy optimization.

Specifically, for each prompt $q$, we sample a group of $G$ responses $\{o_1, o_2, \ldots, o_G\}$ from the old policy $\pi_{\theta_{\mathrm{old}}}$, and compute their normalized group-relative advantages:
\begin{equation}
A_i =
\frac{
r_i-\operatorname{mean}(\{r_1,r_2,\ldots,r_G\})
}{
\operatorname{std}(\{r_1,r_2,\ldots,r_G\})
}.
\end{equation}
The sequence-level importance ratio for response $o_i$ in GSPO is formulated as:
\begin{equation}
s_i(\theta)=
\left(
\frac{\pi_\theta(o_i\mid q)}
{\pi_{\theta_{\mathrm{old}}}(o_i\mid q)}
\right)^{1/|o_i|}=
\exp\left(
\frac{1}{|o_i|}
\sum_{t=1}^{|o_i|}
\log
\frac{
\pi_\theta(o_{i,t}\mid q,o_{i,<t})
}{
\pi_{\theta_{\mathrm{old}}}(o_{i,t}\mid q,o_{i,<t})
}
\right),
\end{equation}
where the length normalization factor $\frac{1}{|o_i|}$ effectively mitigates the variance introduced by varying response lengths. The policy is then optimized using the following sequence-level clipped objective:
\begin{equation}
\begin{split}
    \mathcal{J}_{\mathrm{GSPO}}(\theta) = \mathbb{E}_{\{o_i\}_{i=1}^G \sim \pi_{\theta_{\mathrm{old}}}(\cdot \mid q)} \Bigg[ & \frac{1}{G} \sum_{i=1}^{G} \min\Big( s_i(\theta) A_i, \, \operatorname{clip}(s_i(\theta), 1-\epsilon, 1+\epsilon) A_i \Big) \\
    & - \beta \cdot \mathbb{D}_{\mathrm{KL}}(\pi_\theta \| \pi_{\mathrm{ref}}) \Bigg],
\end{split}
\end{equation}
where $\epsilon$ is the clipping hyperparameter that restricts policy updates within a trust region, $\beta$ adjusts the strength of the KL divergence penalty, and $\pi_{\mathrm{ref}}$ is the reference policy model employed to enhance training stability.

\subsection{Implementation Details}\label{app:implement_details}
In both the IG-SFT and Cap-SFT stages, the model is trained for 2 epochs with a batch size of 64, a learning rate of $2 \times 10^{-5}$, and a load balancing loss coefficient of $1 \times 10^{-3}$. In the IG-RL stage, training is performed for 1 epoch with a batch size of 64 and a learning rate of $5 \times 10^{-6}$. For each query, we sample 8 candidate responses using a temperature of 1.0. The KL-divergence regularization coefficient $\beta$ is set to 0.04 and the clipping hyperparameter $\epsilon$ is set to $1 \times 10^{-3}$. Throughout all training stages, both the video and audio encoders remain frozen, with updates applied exclusively to the adapters and the LLM backbone.

During both training and evaluation, to balance long-video processing efficiency with the the model's context window limitation without sacrificing frame rate and spatial resolution, we uniformly partition long videos into segments with a maximum duration of two minutes, while leaving shorter videos intact. All video inputs are sampled at 2 FPS, with a maximum spatial frame resolution of $256 \times 32 \times 32$ pixels. All training is conducted on 16 NVIDIA H200 GPUs, while evaluation is performed on NVIDIA H20 GPUs.

\section{Details of the Benchmarks}
This section provides detailed descriptions of the benchmarks utilized in our evaluation.

\textbf{CapRiCorn-1K} comprises 1,000 videos covering diverse domains and long temporal horizons, and is designed to evaluate both overall video captioning quality and subject referential consistency across varying durations and scenarios. By leveraging generated captions in downstream understanding and generation tasks, CapRiCorn-1K demonstrates strong correlations between its evaluation results and downstream task performance.

\textbf{UGC-VideoCap} consists of 1,000 short-form TikTok videos, each under 60 seconds and featuring at least one meaningful audio segment of five seconds or more. Caption quality is evaluated by a judge model that assigns scores on a 1-to-5 scale across three dimensions: visual, audio, and details. These dimension scores are then normalized and aggregated to produce a final caption quality score.

\textbf{WorldSense} focuses on scenarios requiring tightly coupled audiovisual understanding, requiring models to effectively leverage synergistic audiovisual perception. It contains 1,662 clips across eight semantic domains, accompanied by 3,172 multiple-choice question-answer pairs spanning 26 diverse tasks. In our experiments, we evaluate the quality of generated captions by feeding them into a judge model and measuring their effectiveness in enabling accurate question answering.

\section{Additional Analysis}
\subsection{Why Not Interleave Identity Gallery Construction with Captioning?}
As introduced in Section~\ref{sec:pigeons}, PIGEONS adopts a two-phase inference procedure: it first scans the entire video to construct a global identity gallery, and then performs segment-wise captioning conditioned on the fixed gallery. A seemingly more elegant alternative would be to interleave these two processes, where the model processes each segment sequentially, updates the identity gallery online, and immediately generates the corresponding caption.

However, a critical limitation of this approach emerges when an entity appears only sparsely in early segments, and thus being held only in the candidate gallery, but appears frequently in later segments. Once the cumulative occurrences justify its promotion to the identity gallery, the previously generated captions for earlier segments cannot be retroactively modified, resulting in fragmented and misaligned subject references across segments.

To address this issue, a potential workaround is to implement a backtracking mechanism. Whenever a candidate entity is promoted to the identity gallery at segment $i$, the model revisits the segment where the entity first appeared and regenerates all captions from that segment to the current one using the updated gallery. We refer to this strategy as \textit{Interleaved Gallery-Captioning}.

To compare the computational efficiency of Interleaved Gallery-Captioning against PIGEONS, we formulate their inference costs as follows. Let $\mathcal{C}_{\mathrm{enc}}$, $\mathcal{C}_{\mathrm{IG}}$, and $\mathcal{C}_{\mathrm{cap}}$ denote the computational costs of encoding a single segment's multimodal input, updating the identity gallery, and generating a caption, respectively. Let $N$ denote the total number of video segments.

\subsubsection{Computational Analysis without Caching}
Without considering feature caching, PIGEONS processes each segment twice: once during the initial identity gallery construction and once during the caption generation phase. Its total inference cost is therefore given by:
\begin{equation}
C_{\mathrm{PIGEONS}} = N \left(2\mathcal{C}_{\mathrm{enc}} + \mathcal{C}_{\mathrm{IG}} + \mathcal{C}_{\mathrm{cap}}\right).
\end{equation}

For Interleaved Gallery-Captioning, the best-case scenario occurs when no candidate entity is ever promoted throughout the entire video. In this case, each segment is processed once, yielding:
\begin{equation}
C_{\mathrm{Interleaved}}^{\mathrm{best}} = N \left(\mathcal{C}_{\mathrm{enc}} + \mathcal{C}_{\mathrm{IG}} + \mathcal{C}_{\mathrm{cap}}\right).
\end{equation}

Under this ideal condition, Interleaved Gallery-Captioning avoids the additional multimodal encoding incurred by the initial gallery construction phase in PIGEONS.

In contrast, consider the worst-case scenario where, after the first segment, every subsequent segment triggers the promotion of an entity that initially appeared in the first segment. When processing segment $i$, the model is forced to regenerate captions for all previous segments. To avoid redundant computation, the caption of the current segment is generated only after backtracking is completed. The total cost is formulated as:
\begin{equation}
\begin{aligned}
C_{\mathrm{Interleaved}}^{\mathrm{worst}} 
&= \left(\mathcal{C}_{\mathrm{enc}} + \mathcal{C}_{\mathrm{IG}} + \mathcal{C}_{\mathrm{cap}}\right) + \sum_{i=2}^{N} \Big[ \left(\mathcal{C}_{\mathrm{enc}} + \mathcal{C}_{\mathrm{IG}}\right) + i \left(\mathcal{C}_{\mathrm{enc}} + \mathcal{C}_{\mathrm{cap}}\right) \Big] \\
&= N \cdot \mathcal{C}_{\mathrm{IG}} + \left(\frac{N(N+3)}{2}-1\right)\mathcal{C}_{\mathrm{enc}} + \frac{N(N+1)}{2}\mathcal{C}_{\mathrm{cap}}.
\end{aligned}
\end{equation}

Compared with PIGEONS, the additional computational cost is:
\begin{equation}
C_{\mathrm{Interleaved}}^{\mathrm{worst}} - C_{\mathrm{PIGEONS}} = \frac{(N-2)(N+1)}{2}\mathcal{C}_{\mathrm{enc}} + \frac{N(N-1)}{2}\mathcal{C}_{\mathrm{cap}},
\end{equation}
which contains a quadratic-order term with respect to the total number of segments $N$, indicating that repeated backtracking becomes prohibitively inefficient for long videos.

We further analyze the expected inference cost. For the $i$-th segment ($i \geq 2$), let $p$ denote the probability that at least one candidate entity is promoted to the identity gallery. For mathematical tractability, we assume that the first appearance index of a promoted entity is uniformly distributed among the preceding segments. If the promoted subject first appeared at segment $j$, where $j \in \{1, \ldots, i-1\}$, the model regenerates captions from the $j$-th to the $i$-th segment, totaling $i-j+1$ segments. Therefore, the expected backtracking length $\mathbb{E}[L_i]$ is:
\begin{equation}
\mathbb{E}[L_i]
=
\frac{1}{i-1}\sum_{j=1}^{i-1}(i-j+1)
=
\frac{i+2}{2}.
\end{equation}

Accordingly, the expected computational cost of processing segment $i$ is:
\begin{equation}
\mathbb{E}[C_i] = \left(\mathcal{C}_{\mathrm{enc}} + \mathcal{C}_{\mathrm{IG}}\right) + (1-p) \cdot \mathcal{C}_{\mathrm{cap}} + p \cdot \mathbb{E}[L_i] \cdot \left(\mathcal{C}_{\mathrm{enc}} + \mathcal{C}_{\mathrm{cap}}\right),
\end{equation}
and the total expected cost becomes:
\begin{equation}
\begin{aligned}
\mathbb{E}[C_{\mathrm{Interleaved}}] 
&= \left(\mathcal{C}_{\mathrm{enc}} + \mathcal{C}_{\mathrm{IG}} + \mathcal{C}_{\mathrm{cap}}\right) + \sum_{i=2}^{N} \mathbb{E}[C_i] \\
&= N\left(\mathcal{C}_{\mathrm{enc}} + \mathcal{C}_{\mathrm{IG}}\right) + \left[N-(N-1)p\right]\mathcal{C}_{\mathrm{cap}} + \frac{p(N^2+5N-6)}{4}\left(\mathcal{C}_{\mathrm{enc}} + \mathcal{C}_{\mathrm{cap}}\right).
\end{aligned}
\end{equation}

Compared with PIGEONS, the expected additional inference cost is:
\begin{equation}
\mathbb{E}[C_{\mathrm{Interleaved}}] - C_{\mathrm{PIGEONS}} = \frac{p(N^2+5N-6)}{4}\left(\mathcal{C}_{\mathrm{enc}} + \mathcal{C}_{\mathrm{cap}}\right) - N \cdot \mathcal{C}_{\mathrm{enc}} - (N-1)p \cdot \mathcal{C}_{\mathrm{cap}}.
\end{equation}

For any constant promotion probability $p>0$, the quadratic term dominates as $N$ increases, making PIGEONS more scalable for long-video captioning.

\subsubsection{Computational Analysis with Representation Caching}
Furthermore, we extend our analysis to a more practical setting where segment-level multimodal representations are cached after their first computation, thereby eliminating repeated feature encoding during backtracking.

For PIGEONS, after its initial scanning of all segments to build the identity gallery, cached segment representations can be directly reused for captioning. Its total inference cost reduces to:
\begin{equation}
C_{\mathrm{PIGEONS}}^{\mathrm{cache}} = N\left(\mathcal{C}_{\mathrm{enc}} + \mathcal{C}_{\mathrm{IG}} + \mathcal{C}_{\mathrm{cap}}\right).
\end{equation}

For Interleaved Gallery-Captioning, representations caching removes repeated input encoding during backtracking. However, caption regeneration remains unavoidable. Therefore, the expected cost of processing segment $i$ becomes:
\begin{equation}
\mathbb{E}[C_i^{\mathrm{cache}}] = \left(\mathcal{C}_{\mathrm{enc}} + \mathcal{C}_{\mathrm{IG}}\right) + (1-p) \cdot \mathcal{C}_{\mathrm{cap}} + p \cdot \mathbb{E}[L_i] \cdot \mathcal{C}_{\mathrm{cap}}.
\end{equation}

Accordingly, the expected total inference cost is:
\begin{equation}
\begin{aligned}
\mathbb{E}[C_{\mathrm{Interleaved}}^{\mathrm{cache}}] 
&= \left(\mathcal{C}_{\mathrm{enc}} + \mathcal{C}_{\mathrm{IG}} + \mathcal{C}_{\mathrm{cap}}\right) + \sum_{i=2}^{N} \mathbb{E}[C_i^{\mathrm{cache}}] \\
&= N\left(\mathcal{C}_{\mathrm{enc}} + \mathcal{C}_{\mathrm{IG}} + \mathcal{C}_{\mathrm{cap}}\right) + \frac{p(N^2+N-2)}{4}\mathcal{C}_{\mathrm{cap}}.
\end{aligned}
\end{equation}

Consequently, even with caching of segment representations, Interleaved Gallery-Captioning still incurs an additional cost of:
\begin{equation}
\mathbb{E}[C_{\mathrm{Interleaved}}^{\mathrm{cache}}] - C_{\mathrm{PIGEONS}}^{\mathrm{cache}} = \frac{p(N^2+N-2)}{4}\mathcal{C}_{\mathrm{cap}},
\end{equation}
which continues to grow quadratically with the number of segments $N$, further demonstrating the superiority of PIGEONS for long-video captioning.

\begin{figure}[t]
    \centering
    \includegraphics[width=\linewidth]{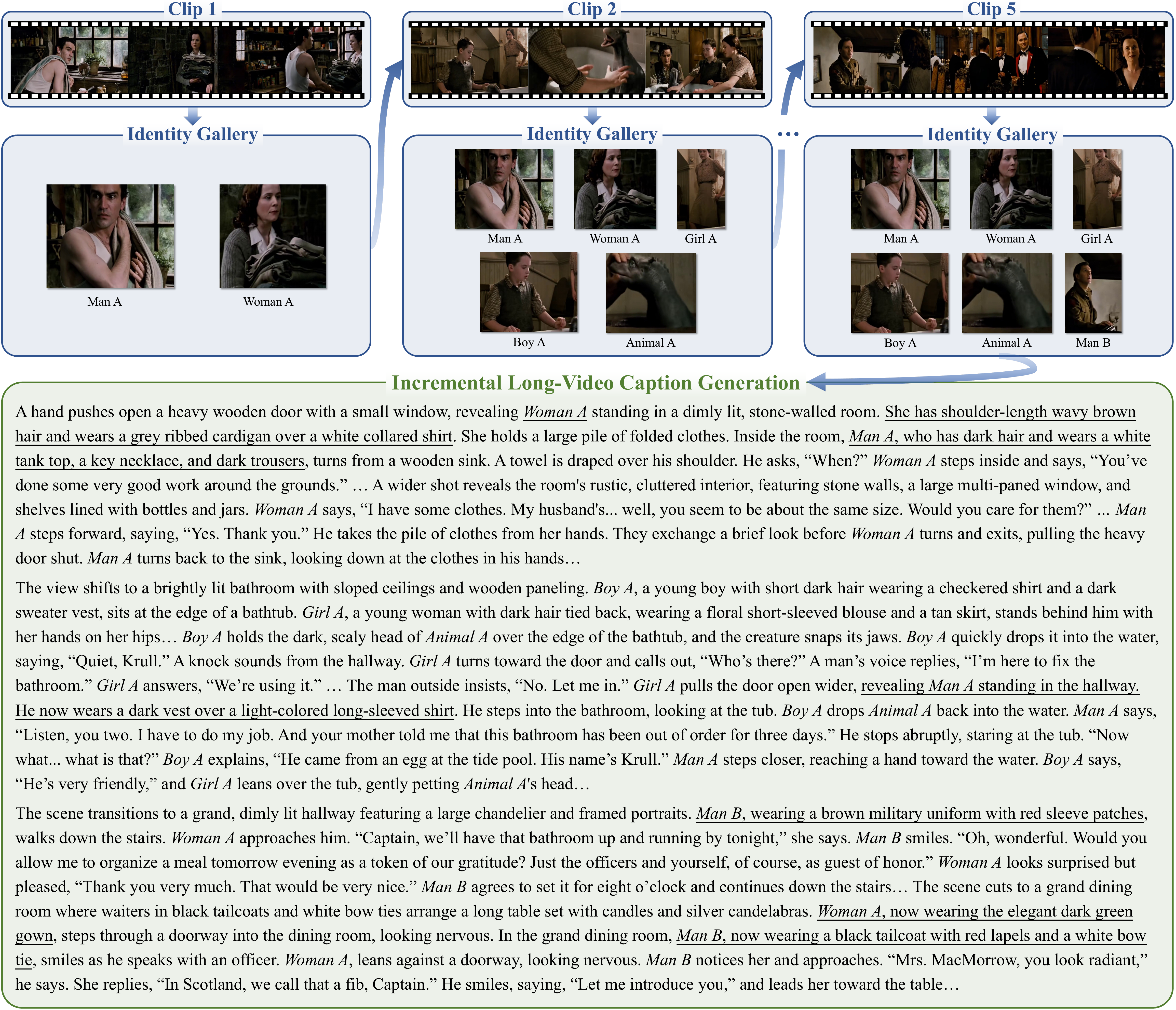}
    \caption{Qualitative captioning results of PIGEONS. Complex cross-segment subject references involving changes in clothing are \underline{underlined}, highlighting PIGEONS's robustness in maintaining accurate subject referential consistency. The overarching narrative summaries are omitted for clarity.}
    \label{fig:case_1}
\end{figure}

\subsection{How Long a Video Can PIGEONS Support?}
Since PIGEONS employs a segment-wise captioning strategy, its inference pipeline is theoretically agnostic to video duration: an arbitrarily long video can be partitioned into shorter clips and processed incrementally. In practice, however, the maximum supported duration is ultimately constrained by the context window of the base model to process individual segments. This limitation primarily manifests in two extreme scenarios: (1) videos characterized by highly complex storyline may involve a multitude of recurring subjects, causing the persistent identity gallery to expand beyond the context budget; and (2) as video length scales continuously, the evolving overarching narrative summary, although substantially more compact than accumulating all preceding captions, may eventually exceed the context window.

Notably, despite being trained on videos with an average duration of only 3.43 minutes, PIGEONS exhibits encouraging temporal generalization. It achieves strong performance on the CapRiCorn-1K and WorldSense benchmarks, reliably handling videos extending up to 10 and 12 minutes, respectively. Therefore, although structurally complex or exceptionally long videos might eventually hit the context ceiling of the current base model, PIGEONS offers a highly scalable approach that helps extend the temporal boundaries of detailed long-video captioning.

\begin{figure}[t]
    \centering
    \includegraphics[width=\linewidth]{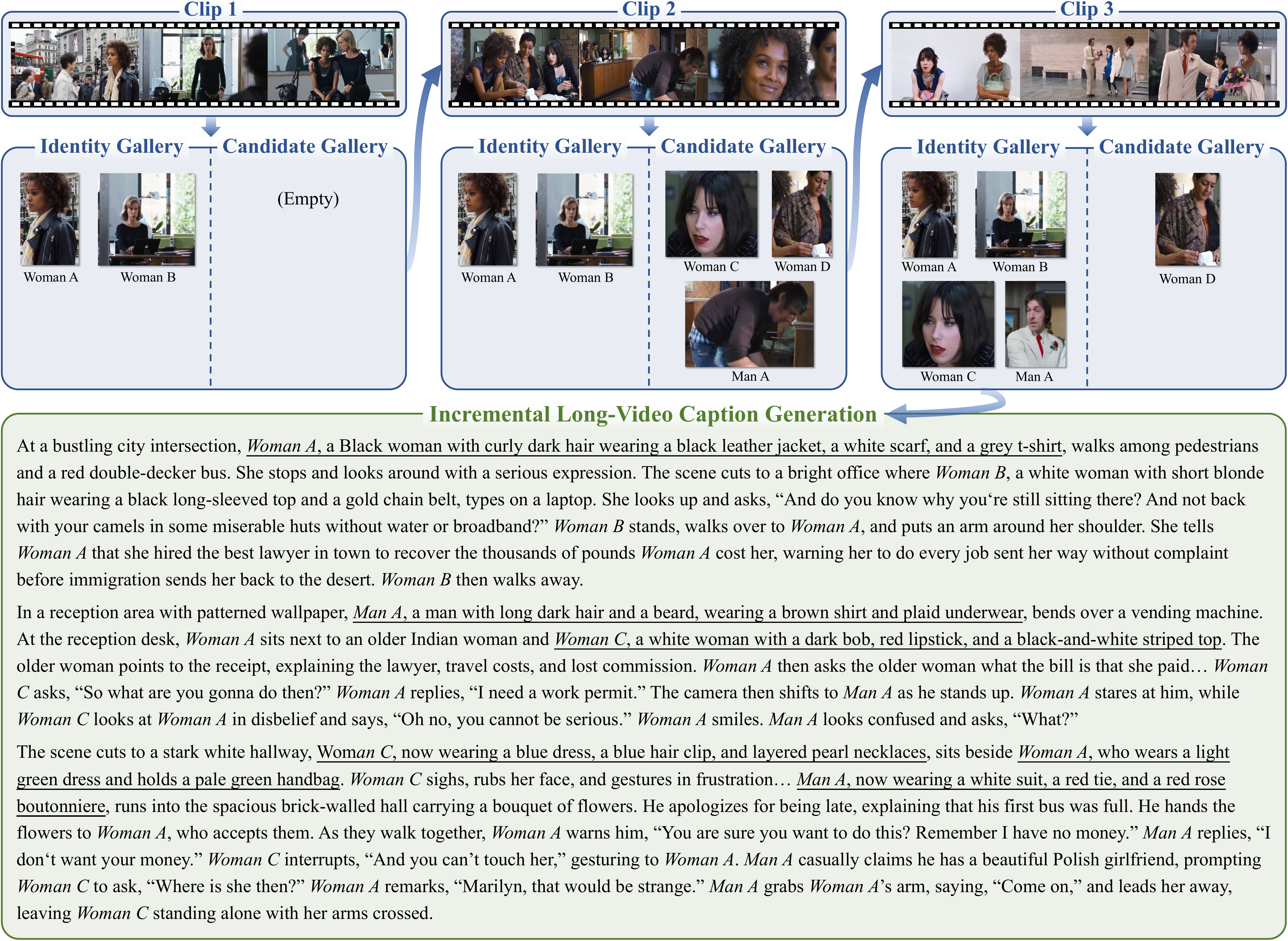}
    \caption{Qualitative captioning results of PIGEONS. Complex cross-segment subject references involving changes in clothing are \underline{underlined}, highlighting PIGEONS's robustness in maintaining accurate subject referential consistency. The overarching narrative summaries are omitted for clarity.}
    \label{fig:case_2}
\end{figure}

\subsection{Why Do Benchmark Scores Increase with Video Duration?}
On the CapRiCorn-1K and WorldSense benchmarks, which feature videos of varying lengths (Tables~\ref{tab:main1} and~\ref{tab:main2}), we observe that PIGEONS generally achieves higher scores on longer videos, except for subject referential consistency. Through qualitative analysis, we find that this trend is partly attributable to differences in annotation granularity across video durations. Specifically, keypoint annotations and evaluation questions for shorter videos often focus on fine-grained visual details, whereas those for longer videos tend to emphasize higher-level semantic events and overall narrative understanding. Therefore, although longer videos contain substantially richer audiovisual content and introduce more complex temporal dependencies, their evaluation objectives and focal points differ, indicating that to some extent, evaluation difficulty may not necessarily scale linearly with video duration. Nevertheless, the performance margin by which PIGEONS outperforms competing baselines progressively widens as video duration increases, demonstrating its strong capability in handling the challenges introduced by long-form videos.

\section{Qualitative Results}
In Figures~\ref{fig:case_1} and~\ref{fig:case_2}, we provide the qualitative long-video captioning results produced by PIGEONS. Benefiting from the persistent identity gallery and evolving overarching narrative summary, PIGEONS accurately and comprehensively captures fine-grained audiovisual details while maintaining robust cross-segment referential consistency for the same subject, even under challenging clothing changes, achieving superior long-range narrative coherence throughout the video.

\section{Details of Prompts}\label{app:prompts}
\subsection{Prompts for Identity Gallery Construction}
Figures~\ref{fig:sbj_ts_prompt} to~\ref{fig:crop_review} present the prompts for generating identity gallery construction data, covering subject and timestamp identification, frame review and refinement, as well as crop review and refinement.

\subsection{Prompts for Caption Generation}
Figure~\ref{fig:caption_gen} illustrates the prompt utilized to generate segment-level caption annotations and update the evolving overarching narrative summary.

\begin{tcolorbox}[colback=white!95!gray, colframe=gray!50!black, rounded corners, enhanced, breakable, title={Prompts for subject and timestamp identification}]
You analyze one clip at a time from a long video and maintain two identity repositories across all clips:

- `subjects`: confirmed primary subjects that meaningfully drive, influence, or shape the video's narrative and are visually re-identifiable.

- `candidates`: secondary or not-yet-confirmed subjects whose events should be tracked in case they later become narratively important, provided there is already concrete evidence that they may matter to the narrative and they are visually re-identifiable as distinct identities.\\

The attached video is the current clip. Any reference images are labeled as `image\_N`, with their mapping supplied in the user message. All timestamps are seconds relative to the start of the current clip, regardless of clip index. For new subjects and candidate upserts, output every `timestamp\_s` value with exactly one decimal place, such as `12.5`.\\

First write a concise natural-language analysis summarizing visible, narratively relevant identities and repository implications in the current clip. Every identity that will appear in the final JSON must be mentioned in this analysis with its `subject\_id`, the visual or narrative evidence supporting the decision. For new visual references, include the selected `timestamp\_s` or timestamp evidence that makes the later JSON decision traceable. For candidate promotions, explain the promotion evidence and state that the saved candidate reference is reused. If no updates are needed, briefly state why no visible identity meets the threshold. Then end the response with one JSON object and no prose after it.\\

\#\# Responsibilities

Make all repository decisions for the current clip:

- Re-identify existing subjects and candidates. Reuse an existing `subject\_id` only when the identity is genuinely the same.

- Add a narratively important new identity directly to `subjects` only when the current clip provides sufficiently distinctive visual evidence to identify and reliably re-identify it.

- Add a secondary identity to `candidates` only when an observed action, interaction, recurrence, or relationship supplies concrete evidence of possible narrative relevance.

- Promote an existing candidate when it has become narratively important by upserting the same ID into `subjects` and including that ID in `candidate\_deletes`.

- Record a concise current-clip `event` for every candidate upsert. It should explain what the candidate does in this clip and help support a future promotion.

- Do not create an upsert solely to re-confirm an existing identity whose saved reference is already sufficiently discriminative. Promotions automatically reuse the existing reference.\\

\#\# Narrative Eligibility

- Treat repository inclusion as a conservative narrative decision, not as detection of every visible person, animal, or object.

- Mere appearance is not narrative significance. Do not add an identity just because it is the first or only visible person, centered, clearly filmed, or easy to identify.

- Background extras, bystanders, incidental visitors, and generic objects should be omitted unless they have a demonstrable narrative effect and the identity is visually re-identifiable.

- `candidates` are not a holding area for every uncertain identity. Do not add a candidate based only on the possibility that it might matter later.

- Do not lower the narrative threshold because repositories have remained empty. Empty repositories are valid and preferable to weak identities.\\

\#\# Identity Rules

- Each `subject\_id` must refer to exactly one identity throughout the video.

- Create a new ID only when visible features are distinctive enough to support reliable re-identification in later clips. Do not create an ID from repeated presence alone.

- For objects, recognizing the category is not enough. The specific physical instance must have stable, distinctive features such as markings, damage, customization, shape, or other persistent cues. Ordinary interchangeable objects should be omitted.

- A generic or partial view that cannot uniquely characterize the identity is insufficient, including isolated limbs, silhouettes, heavily occluded bodies, or generic backs.

- New IDs must use the independent `TYPE\_LETTER` namespace, such as `WOMAN\_A`, `MAN\_B`, `GIRL\_A`, `BOY\_A`, `DOG\_B`, or `VEHICLE\_A`. Use one uppercase suffix letter and no digits, prefixes, angle brackets, or multi-letter suffixes.

- For human subjects, prefer `man`/`woman` for adults and `boy`/`girl` for children. Use `person` only when age and gender cannot be determined reliably.

- Only delete a candidate as part of promotion. Never delete a confirmed subject.\\

\#\# Field Rules

- `subject\_description` must be concise and discriminative. Prefer stable intrinsic appearance, type, and identity cues over temporary pose or action.

- `event` is required and non-empty for every candidate upsert. It should record the candidate's current-clip action or interaction that may support future promotion.

- For new subject upserts and all candidate upserts, `timestamp\_s` must be within the current clip and select a sharp, minimally occluded, sufficiently large full-frame view of the intended identity. A later stage will review the frame and produce the bbox, do not output bbox here.

- For subject upserts that promote an existing candidate, include the same ID in `candidate\_deletes` and omit `timestamp\_s`. The saved candidate reference will be reused, so no new frame extraction is needed.\\

\#\# Output Format

End the response with exactly this JSON shape. The example below shows all three update types at once: a new confirmed subject, a candidate update, and an existing candidate promoted to `subjects`. In a real response, use empty arrays
for sections with no updates. All three top-level keys are required; do not add other keys.

\{
  "subjects\_upsert": [
    \{
      "subject\_id": "WOMAN\_A",
      "subject\_description": "adult woman with short silver hair and round glasses",
      "event": "",
      "timestamp\_s": 12.5
    \},
    \{
      "subject\_id": "DOG\_A",
      "subject\_description": "small white dog with a black patch over its left eye",
      "event": "Runs to the woman and changes the focus of the scene."
    \}
  ],
  "candidates\_upsert": [
    \{
      "subject\_id": "MAN\_A",
      "subject\_description": "adult man in a navy jacket carrying a clipboard",
      "event": "Enters the workshop and briefly speaks with the woman.",
      "timestamp\_s": 22.7
    \}
  ],
  "candidate\_deletes": ["DOG\_A"]
\}

\end{tcolorbox}
\captionof{figure}{Prompts for subject and timestamp identification.}
\label{fig:sbj_ts_prompt}

\begin{tcolorbox}[colback=white!95!gray, colframe=gray!50!black, rounded corners, enhanced, breakable, title={Prompts for frame review, including timestamp refinement and bounding-box proposal}]
You review full video frames for proposed subject reference timestamps, output bboxes immediately for suitable frames, and revise timestamps for unsuitable frames.\\

Each image is a full frame extracted at the proposed timestamp. The attached video is also available only for timestamp correction when a full frame is unsuitable. The user message maps every image to a `detection\_index` and
intended identity.\\

If the frame is suitable, return one tight bounding box for that `detection\_index` immediately. Do not ask for or wait for another image round. Bounding boxes use normalized integer coordinates in exact order `[ymin, xmin, ymax, xmax]`, each value from 0 to 1000. The box should contain
the informative subject instance without cutting off discriminative parts.

If the frame is unsuitable, return one better `timestamp\_s` within the current clip for that `detection\_index` and do not return a bbox for it yet. `timestamp\_s` is in seconds and must use exactly one decimal place, such as `12.5`.\\

Do not change identity, repository membership, events, promotion decisions, or descriptions.\\

First write a concise natural-language analysis explaining what you checked in the frame images, why each suitable frame gets its bbox, and why each unsuitable frame gets a corrected timestamp. The analysis must stay outside the JSON. Then end the response with one JSON object:

\{
  "boxes": [
    {
      "detection\_index": 1,
      "bounding\_box": [120, 210, 920, 760]
    }
  ],
  "timestamp\_revisions": [
    {
      "detection\_index": 2,
      "timestamp\_s": 12.5
    }
  ]
\}

\end{tcolorbox}
\captionof{figure}{Prompts for frame review, including timestamp refinement and bounding-box proposal for suitable frames.}
\label{fig:frame_review}

\begin{tcolorbox}[colback=white!95!gray, colframe=gray!50!black, rounded corners, enhanced, breakable, title={Prompts for crop review and refinement}]
You are the visual verification stage for proposed repository reference crops.\\

You receive the current clip video plus crop images produced from reviewed timestamps and proposed bounding boxes. The user message contains the current repository, the immutable repository patch, and a mapping from image label to `detection\_index`.\\

Review each crop independently:

1. Decide whether it clearly depicts the intended identity and matches its `subject\_description`.

2. Reject crops that are wrong, blurry, too small, badly framed, heavily occluded, or not discriminative enough.

3. Make the final `update\_reference` decision. New accepted identities must update the reference; existing identities should update only when the crop is meaningfully better than the saved reference.

4. If a crop is rejected, use the current clip to provide corrected crop parameters for that same `detection\_index`: a better `timestamp\_s` and normalized `bounding\_box`.\\

Each crop comes from a full frame that was already accepted as suitable for bounding-box annotation. Therefore, for rejected crops, first try to keep the same `timestamp\_s` and fix only the `bounding\_box`. Change `timestamp\_s` only
when the accepted full frame still cannot yield a correct, discriminative crop for the intended identity. Timestamp changes are allowed, but should be rare and explicitly justified in the natural-language analysis.\\

Do not modify repository membership, promotion, events, IDs, timestamps, boxes, or descriptions except through rejected-crop `timestamp\_s` and `bounding\_box` corrections. `timestamp\_s` is in seconds relative to the current clip and must use exactly one decimal place, such as `21.4`. Bounding boxes use normalized integer coordinates in exact order `[ymin, xmin, ymax, xmax]`, each value from 0 to 1000.\\

First write a concise natural-language analysis explaining the crop quality, identity match, reference-update judgment, and any rejected-crop correction for each image. The analysis must stay outside the JSON. Then end the response with one JSON object:

\{
  "reviews": [
    {
      "detection\_index": 1,
      "accepted": true,
      "update\_reference": true,
      "update\_reference\_reason": "This new identity has no saved reference."
    },
    {
      "detection\_index": 2,
      "accepted": false,
      "update\_reference": false,
      "update\_reference\_reason": "Rejected crops cannot update the reference.",
      "timestamp\_s": 21.4,
      "bounding\_box": [100, 240, 940, 790]
    }
  ]
\}

\end{tcolorbox}
\captionof{figure}{Prompts for crop review and refinement.}
\label{fig:crop_review}

\begin{tcolorbox}[colback=white!95!gray, colframe=gray!50!black, rounded corners, enhanced, breakable, title={Prompts for caption and overarching summary generation}]
You are an expert long-video audiovisual captioner. A long video with its audio track has been split into chronological clips. For each current clip, produce:

1. A detailed audiovisual caption describing only the current clip.

2. A compact structured summary of the global narrative information covered through the end of the current clip.\\

The user message provides the current clip, subject reference images with their ID labels, the structured summary through the previous clip, and the previous clip's full caption.

The final long-video caption will be created by directly concatenating all clip captions in chronological order without an additional rewriting pass. Therefore, each caption must work as the next contiguous passage of one coherent document.\\

\#\# Identity rules

- Match visible identities against the labeled reference images. Use an exact stable ID such as `WOMAN\_A` only when the match is visually supported.

- Preserve every supplied ID exactly. Never add any prefix, convert an ID to a numeric or angle-bracket form, rename, normalize, translate, re-letter, or invent a new ID.

- Reference images are identity evidence only. They never replace describing a confirmed subject's visible appearance in the video. When a subject first appears in the concatenated captions, describe their currently visible appearance, including salient clothing, hairstyle, accessories, physical features, and condition. Describe later appearance changes when they occur, such as changed clothing, added or removed accessories, altered hair, makeup, dirt, injuries, wetness, or other visually meaningful changes.

- Describe people, animals, and salient objects that do not have confirmed reference IDs whenever they contribute to the scene. Refer to them with precise generic descriptions and stable descriptive terms within the caption, but never assign them an invented ID.\\

\#\# Current-clip audiovisual caption

- Write the caption so it can be appended directly after the previous clip's caption. Maintain coherent temporal transitions, identity references, terminology, tense, and narrative flow across the clip boundary.

- The concatenated caption must remain understandable on its own. Do not rely on the structured summary, only the captions will appear in the final output.

- Treat both the visual stream and the audio track as primary evidence. The caption must not be a visual-only description when meaningful audio is present.

- Cover every distinct scene and meaningful event in the current clip. Do not summarize only the dominant storyline while skipping brief actions, entrances, exits, background activity or reactions, cutaways, transitions, or meaningful opening and final shots.

- Also describe the meaningful actions, reactions, appearance, and spatial relationships of non-confirmed people, animals, and salient objects. Do not let the presence of confirmed-subject reference images narrow the caption to those subjects alone.

- Describe shot composition and camera behavior when observable and useful: shot scale, angle, viewpoint, framing, focus, camera movement, zooms, cuts, transitions, and meaningful changes in what enters or leaves the frame.

- Integrate audio with the corresponding visual events in the same narrative rather than listing visual and audio information separately. Capture meaningful relationships such as a visible action producing a sound, speech accompanying an action, off-screen sound affecting the scene, or a change in music reinforcing a transition.\\

\#\# Cumulative structured summary

After formulating the caption, update the global narrative information using both the prior structured summary and the current clip.\\

\#\#\# Existing Scenes

- Add or update a scene only after its newly observed information has been expressed in the current caption. The summary must not contain new setting information that is absent from the concatenated captions.

- Each scene must contain exactly two fields: `scene\_id` and `description`.

- Preserve still-valid backgrounds and merge duplicate entries.

- In `description`, capture the setting and any meaningful location, time of day, weather, lighting, atmosphere, layout, or environmental changes.

- Use an empty `scene` list when no background is visually established.\\

\#\#\# Character Relationships

- Add or update a relationship only after it has been explicitly stated in the captions. The summary must not infer relationships from appearance or ambiguous interactions.

- Use the subject IDs from the identity gallery exactly as provided.

- Capture only stable and meaningful relationships between subjects, such as family, friendship, partnership, or other explicitly established social or functional relationships.

- Preserve still-valid relationships, merge duplicate entries, and update an existing relationship when later evidence corrects or materially refines it.

- Use an empty `character\_relationships` list when there is no qualifying relationship.\\

\#\#\# Subject Roles

- Add or update a role only after it has been clearly demonstrated in the captions.

- Use the subject IDs from the identity gallery exactly as provided.

- Capture stable roles or identities that are important for understanding later clips, such as a person's occupation, function, position in a group, or role within the ongoing activity.

- Preserve still-valid roles, merge duplicate entries, and update an existing role when later evidence corrects or materially refines it.

- Use an empty `subject\_roles` list when no qualifying role has been established.\\

\#\#\# Established Facts

- Add or update a fact only after it has been explicitly stated in the captions. The summary must never introduce information absent from the concatenated captions.

- Capture important non-background knowledge that remains valid beyond the moment in which it was introduced and may be needed to understand later clips, such as a project's purpose or defining specification, an experiment's goal or established conclusion, a product's explained mechanism, or an already explained process.

- Preserve still-valid facts, merge duplicates, and update an existing fact when later evidence corrects or materially refines it.

- Use an empty `established\_facts` list when no qualifying fact has been established.
\end{tcolorbox}
\captionof{figure}{Prompts for caption and overarching summary generation.}
\label{fig:caption_gen}

\end{document}

%% file: math_commands.tex
\usepackage{amsmath,amsfonts,bm}

\def\eqref#1{equation~\ref{#1}}

\def\1{\bm{1}}

\DeclareMathAlphabet{\mathsfit}{\encodingdefault}{\sfdefault}{m}{sl}
\SetMathAlphabet{\mathsfit}{bold}{\encodingdefault}{\sfdefault}{bx}{n}

